\documentclass[11pt,letterpaper]{article}
\usepackage{naaclhlt2016}
\usepackage{times}
\usepackage{latexsym}
\usepackage{url}
\usepackage{graphicx}
\usepackage{mathtools}
\usepackage{multirow}
\usepackage{subcaption}
\usepackage{adjustbox}

\naaclfinalcopy 

\title{How would you say that? \\Pictures elicit better NLG data from the crowd.}

\title{Crowd-sourcing NLG Data: Pictures Elicit Better Data.}

\author{Jekaterina Novikova, Oliver Lemon \and Verena Rieser\\
Interaction Lab \\
Heriot-Watt University \\
Edinburgh, EH14 4AS, UK \\
  {\tt \{j.novikova, o.lemon, v.t.rieser\}@hw.ac.uk}}

\date{}

\begin{document}

\maketitle

\begin{abstract}
Recent advances in corpus-based Natural Language Generation (NLG) hold the promise of being easily portable across domains, but require costly training data, consisting of meaning representations (MRs) paired with Natural Language (NL)  utterances. In this work, we propose a novel framework for crowd-sourcing 
 high quality NLG training data, using automatic quality control measures and evaluating different MRs with which to elicit data. 
We show that pictorial MRs result in better NL data being collected than 
 logic-based MRs: 
utterances elicited by pictorial MRs are judged as significantly more natural, more informative, and better phrased, with a significant  increase  in  average quality ratings (around 0.5 points on a 6-point scale), compared to using the logical MRs.
 As the MR becomes more complex, the benefits of  pictorial  stimuli increase. 
 The collected data will be released as part of this submission.

\end{abstract}
\vspace{-0.2cm}
\section{Introduction}
\vspace{-0.2cm}
The overall aim of this research is to develop methods that will allow the full automation of the creation of NLG systems for new applications and domains. Currently deployed technologies for NLG utilise domain-dependent methods including hand-written grammars or domain-specific language templates for surface realisation, both of which are costly to develop and maintain.
Recent corpus-based methods hold the promise of being easily portable across domains, e.g.\ \cite{Angeli2010,Konstas2012,mairesse}, but  require high quality training data consisting of meaning representations (MR) paired with Natural Language (NL) utterances, augmented by alignments between MR elements and NL words.
Recent work \cite{jurcicek:2015:ACL,wen-EtAl:2015:EMNLP} removes the need for alignment, but the question of {\em where to get in-domain training data of sufficient quality}  remains.

In this work, we propose a novel framework for crowd-sourcing 
 high quality NLG training data, using automatic quality control measures and evaluating different meaning representations. So far, we collected 1410 utterances using this framework. The data will be released as part of this submission.
\vspace{-0.2cm}
 \section{Background}
 \vspace{-0.2cm}
Apart from \cite{mairesse2010phrase}, this research is the first to investigate crowdsourcing for collecting NLG data. So far, crowdsourcing is mainly used for
evaluation in the NLG community, e.g.\ \cite{rieser:2014,dethlefs:EMNLP2012}.
Recent efforts in corpus creation via crowdsourcing have proven to be successful in related tasks.
 For example, \cite{callisonburch:2011} showed that crowdsourcing can result in datasets of comparable quality to those created by professional translators given appropriate quality control methods. 
 \cite{mairesse2010phrase} demonstrate that crowd workers can produce NL descriptions from abstract MRs, a method which also has shown success in related NLP tasks, such as Spoken Dialogue Systems \cite{wang2012crowdsourcing} or Semantic Parsing \cite{wang-berant-liang:2015:ACL-IJCNLP}.
However, when collecting
corpora for training NLG systems, new challenges arise:\\
(1) How to ensure the required high quality of the collected data?\\
(2) What types of meaning representations can elicit spontaneous, natural and varied data from crowd-workers?

To address (1), we first filter the crowdsourced data using automatic and manual validation procedures.
We evaluate the quality of crowdsourced NLG data using automatic measures, e.g.\ measuring the semantic similarity of a collected NL utterance.

To address (2), we conduct a principled study regarding the trade-off between semantic expressiveness of the MR and the quality of crowd-sourced utterances elicited for the different semantic representations.
In particular, we investigate translating MRs into pictorial representations as used in, e.g. \cite{black-EtAl:2011:SIGDIAL2011,Williams:2007} for evaluating spoken dialogue systems. We compare these pictorial MRs to text-based MRs used by previous crowd-sourcing work  
 \cite{mairesse2010phrase,wang2012crowdsourcing}. 
 These text-based MRs take the form of Dialogue Acts, such as {\em inform(type[hotel],pricerange[expensive])}.
 However, there is a limit in the semantic complexity that crowd workers can handle~\cite{mairesse2010phrase}. Also, \cite{wang2012crowdsourcing} observed that the semantic formalism unfortunately influences the collected language, i.e.\ crowd-workers are ``primed'' by the words/tokens and ordering used in the MR. 
\vspace{-0.2cm}
\section{Experimental setup} 
\label{exp_setup}
\vspace{-0.2cm}
The experiment was designed to investigate whether we can elicit high-quality Natural Language via crowdsourcing, using different modalities of meaning representation:  textual/logical and pictorial MR. We use the CrowdFlower platform to set up our experiments and to access an online workforce.
\vspace{-0.2cm}
\subsection{Data collection: pictures and text}
\begin{figure}[h]
\centering
\includegraphics[width=0.4\textwidth]{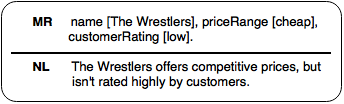}
\caption{Example of an MR-NL pair.} 
\label{fig:pair}
\end{figure}

\vspace{-0.1cm}
The data collected is intended as
training input to a statistical NL generation process, 
but where alignment between words and the MR is left unspecified as in, e.g.\ \cite{jurcicek:2015:ACL,wen-EtAl:2015:EMNLP}. 
The input to the generation process is a pair of MR and NL reference text. Each MR consists of an unordered
 set of \emph{attributes} and  their \emph{values}. The NL reference text  is a Natural Language utterance, possibly consisting of several sentences,  which is provided by a crowd worker for the corresponding MR. An example MR-NL pair is shown in Figure~\ref{fig:pair}.

For the data collection, a set of sixty MRs was prepared, consisting of three, five, or eight attributes and their corresponding values in order to assess different complexities. The eight attributes used in the MRs are shown in Table \ref{tab:attr}.
The order of attributes is randomised so that crowdworkers are not ``primed" by ordering used in the MRs \cite{wang2012crowdsourcing}. 

\begin{table}[htp]
\begin{center}
\begin{adjustbox}{max width=0.48\textwidth}
\begin{tabular}{|l|l|l|}
 \hline
\textbf{Attribute} & \textbf{Data Type} & \textbf{Example value}\\ \hline \hline
name & verbatim string & The Wrestlers, ...\\  \hline
eatType & dictionary & restaurant, pub, ...\\  \hline
familyFriendly & boolean & Yes / No\\  \hline
priceRange & dictionary & cheap, expensive, ...\\  \hline
food & dictionary & Japanese, Italian, ...\\  \hline
near & verbatim string & market square, ... \\ \hline 
area & dictionary & riverside, city centre, ...\\  \hline
customerRating & enumerable & 1 of 5 (low), 4 of 5 (high), ...\\  \hline
\end{tabular}
\end{adjustbox}
\end{center}
\caption{Domain attributes and attribute types.}
\label{tab:attr}
\end{table}%
\vspace{-0.1cm}
75 distinct MRs were prepared in a way that ensures a balance between the number of used attributes in the final dataset. We excluded MRs that do not contain the attribute \emph{name} from the set of MRs with three and five attributes, because we found that such MRs are problematic for crowd workers to create a natural grammatically-correct utterances. For example, crowd workers found it difficult to create an utterance of a high quality based on the MR {\tt priceRange[low], area[riverside], customerRating[low]}.

The textual/logical MRs in our experiment (see Figure \ref{fig:pair}) have the form of a sequence with attributes provided in a random order, separated by commas, and the values of the attributes provided in   square brackets   after each attribute. 

\begin{figure*}[ht]
\centering
\begin{subfigure}{.4\textwidth}
  \centering
  \includegraphics[width=.92\linewidth]{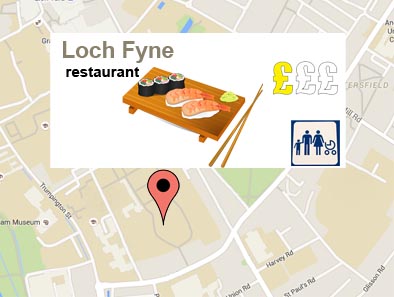}
\end{subfigure}%
\begin{subfigure}{.4\textwidth}
  \centering
  \includegraphics[width=.92\linewidth]{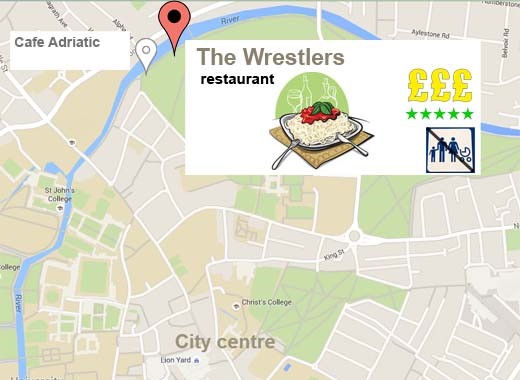}
\end{subfigure}
\caption{Examples of  pictorial MRs. Left: for a family-friendly, Sushi/Japanese restaurant, cheap, neither near the centre of town nor near the  river. Right: for a  restaurant by the river, serving pasta/Italian food, highly rated and expensive, not child-friendly, located near Cafe Adriatic.} 
\label{fig:picture}
\end{figure*}

The pictorial MRs (see Figure \ref{fig:picture}) are semi-automatically generated pictures with a combination of icons corresponding to the appropriate attributes. The icons are located on a background showing a map of a city, thus allowing to represent the meaning of attributes \emph{area} and \emph{near}.

\vspace{-0.1cm}
\subsection{Validation procedures}
\vspace{-0.1cm}
There are several reasons why crowdsourcing might generate poor quality data: 
(1)  The task may be too complex or the instructions might not be clear enough for crowd workers to follow; (2) the financial incentives may be not attractive enough 
 for crowd workers to act conscientiously;  and (3) open-ended job designs without a gold-standard reference test
may allow them to simply randomly click or type ``gibberish" text instead of performing the task. 

In our experiment, we  provided crowd workers with clear and concise instructions of each task. 
 The instructions contained the goal of the data collection, a list of rules describing what is required 
 and what is optional, and three examples of utterances paired with an MR. Instructions for the textual/logical MR and the pictorial MR were intended to be as identical as possible for the two conditions, with only slight unavoidable differences, such as the format of the example MR provided (logical or pictorial).

In terms of financial compensation, crowd workers were paid the standard pay on CrowdFlower, which is
 \$0.02 per page (each containing 1 MR). Workers  were expected to spend about 20 seconds per page. 
 Participants were allowed to complete up to 20 pages, i.e.\ create utterances for up to 20 MRs. Mason and Watts \shortcite{mason2010financial} in their study of financial incentives on Mechanical Turk,  found (counter-intuitively) that increasing the amount of compensation for a particular task does not tend to improve the quality of the results. Furthermore, Callison-Burch and Dredze \shortcite{callison2010creating}  observed that there can be an inverse relationship between the amount of payment and the quality of work, because it may be more tempting for crowd workers to cheat on high-paying tasks if they do not have the skills to complete them. Following these findings, we did not increase the payment for our task over the standard level. 

In order to check for random inputs/``gibberish"  and to control quality of the data, we introduced a
validation procedure, which consisted of two main parts (see sections \ref{prevalid} and \ref{sec:humanEval} for details): 

(1) Automatic pre-validation. The purpose of the automatic validation is to block the submissions of utterances of inappropriate quality. 

(2) Human evaluation of collected data. The purpose of human post-evaluation is to rate the quality of collected utterances.
\vspace{-0.2cm}
\subsection{Automatic Pre-validation}\label{prevalid}
\vspace{-0.1cm}
The first pre-validation step is to select participants that are likely to be native speakers of English. Previous crowdsourcing experiments used different methods to ensure that crowd workers meet this criteria. One option is to create a qualification exam that will screen out non-native speakers. However, as discussed by \cite{sprouse2011validation}, this method is not reliable, as workers can re-take qualification exams multiple times to avoid disqualification. Furthermore, qualification exams severely decrease participation rates, as many crowd workers routinely avoid jobs that require qualification \cite{sprouse2011validation}. Alternatively, Sprouse \shortcite{sprouse2011validation} and Callison-Burch and Dredze \shortcite{callison2010creating} argue for self-identification of the participants, while using their IP addresses to ensure that their geolocation information is consistent with this. In accordance with this, we used   IP addresses  to ensure that participants are located in one of three English-speaking countries - Canada, the United Kingdom, or the United States. In addition, both in the name of the task and in the instructions, we included a requirement that ``Participants must be native speakers of British or American English".

The second pre-validation step checks whether participants spend at least 20 seconds to complete a page of work. This is a standard CrowdFlower option to control the quality of contributions, and it ensures that the contributor is removed from the job if they complete the task too fast.

As a final pre-validation step, we created four JavaScript validators to ensure the submitted utterances are well formed English sentences: 

(1) The first validator  checked if the ready-to-submit utterance only contains legal characters, i.e.\ letters, numbers and symbols ``, . : ;\pounds'". 

(2) The second validator checked whether the length of the utterance (in characters) is not smaller than the required minimal length. The required minimal length was calculated as follows: 
\vspace{-0.2cm}
\begin{equation}
\begin{aligned}
min.length = length.of.MR - \\ number.of.attributes.in.MR \times 10;
\end{aligned}
\end{equation}
Here, {\it length.of.MR} is the total number of characters in the provided MR. {\it Number.of.attributes.in.MR} is either 3, 5 or 8 depending on the number of attributes in the provided MR. {\it 10} is an average length of an attribute name, including two associated square brackets. Thus, $min.length$ is simply an approximation of the total number of characters used for attribute values in each specific MR. 

(3) The third validator checked whether the ready-to-submit utterance contained all the required elements, e.g.\ the name of the described venue or the name of the venue near the described one.

(4) The last validator checked that participants do not submit the same utterance several times.

The automatic validators were tested on the data collected during a pilot test phase and were able to correctly identify and reject 100\% of bad submissions.
\vspace{-0.1cm}
\subsection{Human evaluation of collected data}\label{sec:humanEval}
\vspace{-0.1cm}
While automatic validators help reject some invalid cases, human feedback is needed to assess the quality of the collected data. In a 2nd phase we  evaluated the collected data through a large-scale subjective rating experiment using the CrowdFlower system.

6-point Likert scales were used to collect judgements on the   data, via the following criteria:
\vspace{-0.15cm}
\begin{enumerate} 
\item  {\it Informativeness}. Q1: ``Is this utterance informative? (i.e.\ do you think it provides enough useful information about the venue?)"
\vspace{-0.25cm}
\item {\it Naturalness}. Q2: ``Is this utterance natural? (e.g.\ could it have been produced by a native speaker?)"
\vspace{-0.25cm}
\item  {\it Phrasing}. Q3: ``Is this utterance well phrased? (i.e.\ do you like how it is expressed?)"
\end{enumerate}
\vspace{-0.2cm}
Finally, crowd workers were asked to judge whether the utterance is grammatically correct.
\vspace{-0.15cm}
\section{Results: Collected data}
\vspace{-0.1cm}
In order to maintain a balanced workload distribution between the two MR conditions, we divided the workload into two batches: each batch was posted in the morning of two different workdays. Such a workload distribution was previously described in \cite{wang2012crowdsourcing} as appropriate for a between-subject design. Each batch corresponded to one of two conditions: the first batch contained only textual/logical MRs, and the second one used only pictorial MRs. The analysis presented in the following sections is based on this experimental design.

435 tasks were completed by 134 crowd workers: 70 crowd workers completed 212 tasks based on textual/logical MRs, and 64 crowd workers completed 223 tasks on pictorial MRs. This resulted in collecting 1410 utterances, 744 on textual, and 666 on pictorial MRs. 13 crowd workers   completed the tasks on both types of MR. The utterances created by these 13 subjects for the pictorial MRs were excluded from the analysis, so that it would not violate a between-subject experimental design with a possible learning bias. The final dataset therefore contained  {\em 744 utterances elicited using the textual MRs and 498 utterances elicited using the pictorial MRs}, with 1133 distinct utterances. The dataset will be released with this submission. 

We now use objective measures to assess the effect of the MR modality on the collected NL text. 
\vspace{-0.2cm}
\subsection{Time taken to collect data}
\vspace{-0.1cm}
The data collection for the first batch (only textual/logical MRs) was completed in about 26 hours, while the second one (only pictorial MRs) was completed in less than 18 hours. 

The average duration per task was 352 sec for the pictorial MR, and 347 sec for the textual/logical method, as shown in Table~\ref{tab:data-collected}. A two-way ANOVA was conducted to examine the effect of MR modality and the number of attributes on average task duration. The difference between two modalities was not significant, with $p = 0.76$. There was no statistically significant interaction between the effects  of modality and the number of attributes in the MR, on time taken to collect the data. A   main effects analysis showed that the average duration of utterance creation was significantly longer for larger numbers of attributes, F(2,1236) = 24.99, $p <0.001$, as expected. 

\begin{table}[t]
\small
\centering
\begin{adjustbox}{max width=0.42\textwidth}
\begin{tabular}{|l|c|c|c|c|}
\hline
\multirow{2}{*}{} & \multicolumn{2}{c|}{\textbf{Textual MR}} & \multicolumn{2}{c|}{\textbf{Pictorial MR}} \\ \cline{2-5} 
 & Mean & StDev & Mean & StDev \\ \hline \hline
\textit{Time, sec} & \textit{347.18} & \textit{301.74} & \textit{352.05} & \textit{249.34} \\ 
\multicolumn{1}{|r|}{3 attributes} & 283.37 & 265.82 & 298.97 & 272.44 \\ 
\multicolumn{1}{|r|}{5 attributes} & 321.75 & 290.89 & 355.56 & 244.57 \\ 
\multicolumn{1}{|r|}{8 attributes} & 433.41 & 325.04 & 405.56 & 215.43 \\  \hline
\textit{Length, char} & \textit{100.83} & \textit{46.40} & \textit{93.06} & \textit{37.78} \\ 
\multicolumn{1}{|r|}{3 attributes} & 61.25 & 19.44 & 67.98 & 22.30 \\ 
\multicolumn{1}{|r|}{5 attributes} & 95.18 & 26.71 & 91.13 & 21.19 \\ 
\multicolumn{1}{|r|}{8 attributes} & 144.79 & 41.84 & 121.94 & 40.13 \\ \hline
\textit{No of sentences} & \textit{1.43} & \textit{0.69} & \textit{1.31} & \textit{0.54} \\ 
\multicolumn{1}{|r|}{3 attributes} & 1.06 & 0.24 & 1.07 & 0.25 \\ 
\multicolumn{1}{|r|}{5 attributes} & 1.37 & 0.51 & 1.25 & 0.49 \\ 
\multicolumn{1}{|r|}{8 attributes} & 1.84 & 0.88 & 1.63 & 0.64 \\  \hline

\end{tabular}
\end{adjustbox}
\caption{\label{tab:data-collected} Nature of the data collected with each MR. Italics denote   averages across all numbers of attributes. }
\end{table}
\vspace{-0.2cm}
\subsection{Average length of utterance (characters)}
\vspace{-0.1cm}
The length of collected utterances was calculated as a total number of characters in the utterance, including punctuation. 

The average length of utterance was 101 characters for the textual/logical MR, and 93 characters for the pictorial method, as shown in Table~\ref{tab:data-collected}. A two-way ANOVA was conducted to examine the effect of MR modality and the number of attributes on the length of utterance. There was a statistically significant interaction between the effects  of modality and the number of attributes in the MR, F(2,1236) = 23.74, $p < 0.001$. A main effects analysis showed that the average length of utterance was significantly larger not only for a larger number of attributes, with $p < 0.001$, but also for the utterances created based on a textual/logical MR which had a higher number of attributes, $p < 0.001$. 
\vspace{-0.1cm}
\subsection{Average number of sentences per utterance}
\vspace{-0.1cm}
The task allowed crowd workers to create not only single sentences, but also multi-sentence utterances for any provided MR. 

The average number of sentences per utterance was 1.43 for the textual/logical MR, and 1.31 for the pictorial method, as shown in Table~\ref{tab:data-collected}. A two-way ANOVA was conducted to examine the effect of MR modality and the number of attributes on the number of sentences per utterance. There was a statistically significant interaction between the effects  of modality and the number of attributes in the MR, F(2,1236) = 3.83, $p < 0.05$. A main effects analysis showed that the average number of sentences was significantly larger not only for a larger number of attributes, with $p < 0.001$, but also for the utterances created based on a textual/logical MR which had a higher number of attributes, $p < 0.001$. 
\vspace{-0.1cm}
\subsection{Semantic similarity}
\vspace{-0.1cm}
We now examine the \emph{semantic similarity} of the collected sentences. The concept of semantic similarity aims to measure how well the collected utterances cover the 
meaning provided in the MRs. This concept is similar to that of  Informativeness (see section~\ref{sec:inform}), as a higher value for semantic 
similarity
 shows that more information, originally provided in the MR, was expressed in the NL utterance. However, these two concepts are not interchangeable, as we will explain later in Section~\ref{subs:self-crowd}.  

We used a semi-automatic labelling process to assess the semantics of the collected data and compared them to the given MRs. We first performed spell-checking by using Microsoft Word. Overall, about 7\% of the collected utterances contained one or more spelling errors. 
Note that this result is in line with \cite{wang2012crowdsourcing}, who report 8\% spelling errors for crowd-sourced utterances. 
We  corrected these  by hand.
Next, we used an automated process to assess whether the collected data covers all required semantic concepts in the MR, using text similarity. In particular, we 
calculated a similarity score between the provided MR and the collected utterance, using the UMBC Semantic Text Similarity measure provided by \cite{han2013umbc}, which ranked top in the *SEM 2013 Semantic Textual Similarity shared task. 
This measure is based on distributional similarity and Latent Semantic Analysis (LSA), and is further complemented with semantic relations extracted from WordNet. The score was calculated using a Web API\footnote{http://swoogle.umbc.edu/SimService/api.html} to query the UMBC Semantic Similarity service. 
 
We find that textual MRs elicit text which is significantly more similar to the underlying MR than using pictures (similarity score of 0.62 for pictures vs.\ 0.82 for text, $p<0.005$, where 1 indicates perfect overlap). We attribute this difference to the fact that utterances in response to pictorial MRs are more varied and thus receive lower scores.
For example, the similarity score between``cheap" (in MR) and ``cheap" (in a corresponding utterance) is 1, whereas the similarity between ``cheap" and ``low price" is 0.36 using the UMBC Semantic Text Similarity measure.

As a next step, we normalised the results of semantic similarity on a 1-6 scale,  
 in order to make the results comparable to the human ratings on 6-point Likert scales 
 and compare semantic 
similarity
 to the self-evaluation results. In order to make results comparable, we labelled the semantic similarity of a corresponding utterance as \emph{higher than average} if the result was higher than 4 (53\% of all collected utterances), \emph{lower than average} if the result was lower than 3 (4.3\% of all collected utterances), and \emph{average} otherwise (43\% of all the utterances).
This metric is then used to automatically assess the amount of relevant information from the MR which is preserved in the corresponding NL text, see section \ref{subs:self-crowd}. 



\vspace{-0.1cm}
\section{Results: human evaluation of the collected data}
\vspace{-0.1cm}

While automated or semi-automated metrics provide some useful information about the collected utterances, human feedback is necessary to properly assess their quality. 
In this section, we first compare the   data collected using self-evaluation and crowd evaluation  methods, and later we analyse Informativeness, Naturalness, and Phrasing of the collected utterances. We mostly use parametric statistical methods in our analysis. It has been debated for over 50 years whether Likert-type measurement scales should be analysed using parametric or non-parametric statistical methods \cite{carifio2008resolving}. The use of parametric statistics, however, was justified repeatedly by \cite{carifio2008resolving}, \cite{norman2010likert} and more recently by \cite{murray2013likert} as a ``perfectly
appropriate" \cite{carifio2008resolving} statistical method for Likert scales that may be used by researchers ``with no fear of coming to the wrong conclusion" \cite{norman2010likert}. We therefore present and analyse mean averages (rather than the mode) for the collected judgements.

\vspace{-0.2cm}
\subsection{Self-evaluation vs.\ Crowd evaluation}
\label{subs:self-crowd}
\vspace{-0.2cm}

In our experiment we used two methods to evaluate the quality of collected utterances: self-evaluation and an independent crowd-based evaluation. During the self-evaluation, crowd workers were asked to rank their own  utterances. Note   that   data collected using the self-evaluation method was not intended to    allow us to compare the quality of utterances elicited via pictorial and textual MRs. Rather, this data was collected in order to understand whether self-evaluation  may be a reliable technique to evaluate the quality of created utterances in future studies. 

In the self-evaluation, for each of their own  NL utterances, crowd workers could select either \emph{higher than average, average}, or \emph{lower than average} values for  Informativeness, Naturalness, and Phrasing. 

For  the independent crowd evaluation, a new CrowdFlower task was created.
In this task, crowd workers were asked to look at one utterance at a time and to rate each utterance
using the same procedure.

In order to compare the results of self-evaluation with the results of the independent crowd evaluation, we   labelled the results of perceived Informativeness, Naturalness and Phrasing as \emph{higher than average, average} and \emph{lower than average} in both modes. 

Cohen's kappa coefficient was used to measure inter-rater agreement between the two groups of evaluators, i.e.\ self-evaluators and independent crowd evaluators. 
The statistics did not reveal a significant level of agreement between the two groups of evaluators neither for the scores of Informativeness 
($\kappa$ = 0.014, $p$ = 0.36), nor Phrasing ($\kappa$ = 0.007, $p$ = 0.64), nor Naturalness ($\kappa$ = -0.007, $p$ = 0.62). 

The lack of agreement with the independent evaluation already indicates a potential problem with the self-evaluation method. However, in  order to further assess which group was more reliable in evaluating utterances, we compared their Informativeness scores with the Semantic Similarity 
score of the corresponding utterances. As  discussed before, the concepts of Informativeness and Semantic Similarity
 are  similar to each other, so better agreement between these scores
indicates higher reliability of evaluation results. In particular,  utterances with high Semantic Similarity
 would be expected to have high ratings for Informativeness, as they express more of the concepts from the original MR.

 The percentage agreement between the Informativeness and Semantic Similarity 
was 31.1\%, while for the utterances evaluated independently by the crowd it was 60.3\%. The differences in percentage agreements for the utterances with \emph{good} semantic similarity
 was even higher:  32.1\% for self-evaluators vs.\ 75.1\% for crowd evaluators. This strongly suggests that the evaluation quality of self-evaluators is less reliable than that of the crowd. Therefore, we focus on  the data collected from crowd evaluation for the analysis presented in the following sections.

\vspace{-0.1cm}
\subsection{Informativeness}
\label{sec:inform}
\vspace{-0.1cm}

Informativeness was defined (on the questionnaires) as whether the utterance ``provides enough useful information about the venue". Also see section \ref{sec:humanEval}.
The average score for Informativeness was 4.28 for the textual/logical MR, and 4.51 for the pictorial method, as shown in Table~\ref{tab:evaluation}. A two-way ANOVA was conducted to examine the effect of MR modality and the number of attributes on the perceived Informativeness. There was no statistically significant interaction between the effects  of modality and the number of attributes in the MR, F(2,1236) = 1.79, $p = 0.17$. A main effects analysis showed that the average Informativeness of utterances elicited through the pictorial method (4.51) was significantly higher than that of utterances elicited using the textual/logical modality (4.28), with $p < 0.01$. This is an increase of 0.23 points on the 6-point scale (=4.6\%) in average Informativeness rating for the pictorial condition.

As expected, Informativeness increases with the number of attributes in the MR, in both conditions.

\vspace{-0.1cm}
\subsection{Naturalness}
\vspace{-0.1cm}

Naturalness was defined (on the questionnaires) as whether the utterance ``could have been produced by a native speaker".
The average score for Naturalness was 4.09 for the textual/logical MRs, and 4.43 for the pictorial method, as shown in Table~\ref{tab:evaluation}. A two-way ANOVA was conducted to examine the effects of MR modality and the number of attributes on the perceived Naturalness. There was no statistically significant interaction between the effects  of modality and the number of attributes in the MR, F(2,1236) = 0.73, $p = 0.48$. A main effects analysis showed that the average Naturalness of utterances elicited using the  pictorial modality (4.43)  was significantly higher than that of utterances elicited using the textual/logical modality (4.09), with $p < 0.001$. This is an increase of about 0.34 points on the scale (=6.8\%) for average Naturalness rating for the pictorial condition.

\vspace{-0.1cm}
\subsection{Phrasing}
\vspace{-0.1cm}

Phrasing was defined   as whether   utterances are formulated 
 in a way that  the judges perceived as good English 
 (see section \ref{sec:humanEval}).
The average score for Phrasing was 4.01 for the textual/logical MR, and 4.40 for the pictorial method, as shown in Table~\ref{tab:evaluation}. A two-way ANOVA was conducted to examine the effect of MR modality and the number of attributes on the perceived Phrasing. There was no statistically significant interaction between the effects  of modality and the number of attributes in MR, F(2,1236) = 0.85, $p = 0.43$. A main effects analysis showed that the average Phrasing score for the utterances elicited using the pictorial modality was significantly higher than that of the utterances elicited using the textual/logical modality, with $p < 0.001$. This is an increase of +0.39 points (about 7.8\%) in average Phrasing rating for the pictorial condition.

As the complexity of the MR increases (i.e.\ number of attributes) we note that the pictorial MR outperforms the textual MR still further, with an 11\% boost in Phrasing ratings (+0.55  -- from 3.98 to 4.53 on a 6-point scale -- for 8 attributes) and a similar 9.6\% (+0.48)  increase for Naturalness ratings.

A Pearson product-moment correlation method was used to determine a strong correlation ($r = 0.84, p < 0.001$)
 between Naturalness and Phrasing, suggesting that evaluators treat these concepts as very similar. However, these concepts are not identical, as the evaluation results show. 

\begin{table}[t]
\small
\centering
\begin{adjustbox}{max width=0.42\textwidth}
\begin{tabular}{|l|c|c|c|c|}
\hline
\multirow{2}{*}{} & \multicolumn{2}{c|}{\textbf{Textual MR}} & \multicolumn{2}{c|}{\textbf{Pictorial MR}} \\ \cline{2-5} 
 & Mean & StDev & Mean & StDev \\ \hline \hline
\textit{Informativeness} & \textit{4.28**} & \textit{1.54} & \textit{4.51**} & \textit{1.37} \\
\multicolumn{1}{|r|}{3 attributes} & 4.02 & 1.39 & 4.11 & 1.32 \\ 
\multicolumn{1}{|r|}{5 attributes} & 4.31 & 1.54 & 4.46 & 1.36 \\ 
\multicolumn{1}{|r|}{8 attributes} & 4.52 & 1.65 & 4.98 & 1.29 \\  \hline

\textit{Naturalness} & \textit{4.09***} & \textit{1.56} & \textit{4.43***} & \textit{1.35} \\
\multicolumn{1}{|r|}{3 attributes} & 4.13 & 1.47 & 4.35 & 1.29 \\ 
\multicolumn{1}{|r|}{5 attributes} & 4.07 & 1.56 & 4.41 & 1.36 \\ 
\multicolumn{1}{|r|}{8 attributes} & 4.07 & 1.65 & 4.55 & 1.42 \\  \hline

\textit{Phrasing} & \textit{4.01***} & \textit{1.69} & \textit{4.40***} & \textit{1.52} \\
\multicolumn{1}{|r|}{3 attributes} & 4.01 & 1.62 & 4.37 & 1.47 \\ 
\multicolumn{1}{|r|}{5 attributes} & 4.04 & 1.70 & 4.28 & 1.57 \\ 
\multicolumn{1}{|r|}{8 attributes} & 3.98 & 1.75 & 4.53 & 1.54 \\  \hline

\end{tabular}
\end{adjustbox}
\caption{\label{tab:evaluation} Human evaluation of the data collected with each MR (** = $p<0.01$ and *** = $p<0.001$ for Pictorial versus Textual conditions). Italics denote   averages across all numbers of attributes. }
\end{table}
\vspace{-0.1cm}
 \section{Discussion}\label{discussion}
We have shown that pictorial MRs have specific benefits for elicitation of NLG data from crowd-workers. This may be because, with pictures,  data-providers are not primed by lexical tokens in the MRs, resulting in more spontaneous and natural language, with more variability. For example, rather than seeing {\it child-friendly[yes]} in a logical/textual MR, and then being inclined to say ``It is child-friendly",   crowd-workers who see  an icon representing a child seem  more likely to use a variety of phrases, such as ``good for kids''.
As a concrete example of this phenomenon, from the collected data, consider the picture on the left of figure \ref{fig:picture}, which corresponds to the logical MR: {\it name [Loch Fyne], eatType [restaurant], familyFriendly [yes], priceRange [cheap], food [Japanese]}.\\
The logical MR elicited   utterances such as ``Loch Fyne is a family friendly restaurant serving cheap Japanese food'' whereas the pictorial MR elicited e.g.\ ``Serving low cost Japanese style cuisine, Loch Fyne caters for everyone, including families with small children.''

Pictorial stimuli have also been used in other, related NLP tasks. For example in crowd-sourced evaluations of dialogue systems, e.g.\ \cite{black-EtAl:2011:SIGDIAL2011,Williams:2007}. However, no analysis  was performed regarding the suitability of such representations. 
In \cite{Williams:2007}, for example,
pictures were used to set dialogue goals for users (e.g.\ to find an  expensive Italian restaurant in the town centre).
 This experiment therefore also has a bearing on the whole issue of human NL responses to pictorial task stimuli, and shows for example that pictorial task presentations can elicit more natural variability in user inputs to a dialogue system. Pictorial method can also scale up to more than just single-entity descriptions, e.g. it is possible to show on a map several different pictures representing different restaurants, thus eliciting comparisons.
 Of course, there is a limit in the meaning complexity that pictures can express. 
\vspace{-0.2cm}
\section{Conclusions and Future Work} \label{conclusions}
\vspace{-0.2cm}
We have shown that it is possible to rapidly create high quality NLG data sets for training novel corpus-based Machine Learning methods using crowdsourcing. This now forges the path towards rapidly creating NLG systems for new domains. 
We first show that self-evaluation by crowd workers, of their own provided data, does not agree with an independent crowd-based evaluation, and also that their Informativeness judgements do not agree  with an objective metric of semantic similarity.
We then demonstrate  that pictures elicit better data -- that is, judged by independent evaluators as significantly more informative, more natural, and better-phrased --  than logic-based Meaning Representations. There is no significant difference in the amount of time needed to collect the data, but pictorial representations lead to significantly increased scores for these metrics (e.g.\ of around 0.5 on a 6-point Likert scale).
An error analysis shows that pictorial MRs result in more spontaneous, natural and varied utterances. 
We have done this by proposing a new crowdsourcing framework, where we   introduce an initial automatic validation procedure, which was able to reject 100\% of bad submissions.  The collected data  will be released as part of this submission.

In future work, we will use the collected data to test and further develop corpus-based NLG methods, using Imitation Learning. This technique   promises to be able to learn NLG strategies automatically from unaligned data, similar to  recent work by \cite{jurcicek:2015:ACL,wen-EtAl:2015:EMNLP}.  

\footnotesize
\section*{Acknowledgments}
\vspace{-0.2cm}
This research received funding from the EPSRC projects GUI (EP/L026775/1) and DILiGENt (EP/M005429/1). 
\normalsize

\vspace{-0.3cm}
\bibliography{naaclhlt2016}
\bibliographystyle{naaclhlt2016}

\end{document}